\documentclass[a4paper,fleqn]{cas}

\usepackage[authoryear,longnamesfirst]{natbib}
\usepackage{float}
\usepackage{placeins}
\usepackage{color}

\def\tsc#1{\csdef{#1}{\textsc{\lowercase{#1}}\xspace}}
\tsc{WGM}
\tsc{QE}
\tsc{EP}
\tsc{PMS}
\tsc{BEC}
\tsc{DE}
\usepackage{array} % for fixed width columns
\usepackage{graphicx} % for resizebox
\usepackage{amssymb} % for checkmark
\usepackage{pdflscape} % for rotating the page
\usepackage{rotating} % for rotating tables

\usepackage{paralist}
\usepackage{glossaries}

\glsdisablehyper

\begin{document}
\let\WriteBookmarks\relax
\def\floatpagepagefraction{1}
\def\textpagefraction{.001}

\newacronym{llm}{LLMs}{Large Language Models}
\newacronym{rlhf}{RLHF}{Reinforcement Learning from Human Feedback}
\newacronym{lm}{LM}{Language Modeling}
\newacronym{ai}{AI}{Artificial Intelligence}
\newacronym{sft}{SFT}{Supervised Fine-Tuning}
\newacronym{rm}{RM}{Reward Modeling}
\newacronym{rl}{RL}{Reinforcement Learning}
\newacronym{ppo}{PPO}{Proximal Policy Optimization}

\shorttitle{Effects of RLHF on LLM-Generated Text Detection}

\shortauthors{Xu \& Zubiaga}

\title [mode = title]{Understanding the Effects of RLHF on the Quality and Detectability of LLM-Generated Texts}                      
% \tnotemark[1,2]

%\tnotetext[1]{This document is the results of the research
%   project funded by the National Science Foundation.}

%\tnotetext[2]{The second title footnote which is a longer text matter
%   to fill through the whole text width and overflow into
%   another line in the footnotes area of the first page.}

\author[1]{Beining Xu}[orcid=0009-0000-3342-1782]
\cormark[1]


\ead{b.xu@se23.qmul.ac.uk}
%\ead[url]{www.cvr.cc, cvr@sayahna.org}

\credit{Investigation, Conceptualization, Data Curation, Formal Analysis, Writing - Original Draft}

\affiliation[1]{organization={Queen Mary University of London},
    addressline={Mile End Road}, 
    city={London},
    % citysep={}, % Uncomment if no comma needed between city and postcode
    postcode={E1 4NS}, 
    country={United Kingdom}}

\author[1]{Arkaitz Zubiaga}[orcid=0000-0003-4583-3623]

%\author[2,3]{CV Rajagopal}[%
%   role=Co-ordinator,
%   suffix=Jr,
%\ead{cvr3@sayahna.org}
%\ead[URL]{www.sayahna.org}

\credit{Supervision, Writing - Review \& Editing}

%\affiliation[2]{organization={Sayahna Foundation},
%    city={Jagathy},
    % citysep={}, % Uncomment if no comma needed between city and postcode
 %   postcode={695014}, 
  %  state={Trivandrum},
   % country={India}}

%\author%
%[1,3]
%{Rishi T.}
%\cormark[2]
%\fnmark[1,3]
%\ead{rishi@stmdocs.in}
%\ead[URL]{www.stmdocs.in}

%\affiliation[3]{organization={STM Document Engineering Pvt Ltd.},
 %   addressline={Mepukada}, 
  %  city={Malayinkil},
    % citysep={}, % Uncomment if no comma needed between city and postcode
   % postcode={695571}, 
    %state={Trivandrum},
    %country={India}}

\cortext[cor1]{Corresponding author}
%\cortext[cor2]{Principal corresponding author}

%\fntext[fn1]{This is the first author footnote. but is common to third
 % author as well.}
%\fntext[fn2]{Another author footnote, this is a very long footnote and
%  it should be a really long footnote. But this footnote is not yet
%  sufficiently long enough to make two lines of footnote text.}

%\nonumnote{This note has no numbers. In this work we demonstrate $a_b$
%  the formation Y\_1 of a new type of polariton on the interface
%  between a cuprous oxide slab and a polystyrene micro-sphere placed
%  on the slab.
%  }

\begin{abstract}
 Large Language Models (LLMs) have demonstrated exceptional performance on a range of downstream NLP tasks by generating text that closely resembles human writing. However, the ease of achieving this similarity raises concerns from potential malicious uses at scale by bad actors, as LLM-generated text becomes increasingly difficult to discern from human text. Although detection methods have been developed to address this issue, bad actors can further manipulate LLM-generated texts to make them less detectable. In this work, we study how further editing texts with Reinforcement Learning from Human Feedback (RLHF), which aligns model outputs with human preferences, affects (a) the quality of generated texts for two tasks, and (b) the performance of LLM-generated text detectors, looking at both training-based and zero-shot detection methods. Although RLHF improves the quality of LLM-generated texts, we find that it also tends to produce more detectable, lengthy, and repetitive outputs. Additionally, we observe that training-based detectors are vulnerable to short texts and to texts that incorporate code, whereas zero-shot detectors exhibit greater robustness.
\end{abstract}

% % Research highlights
% \begin{highlights}
%  \item We study the impact of RLHF on LLM texts focused on their quality and detectability.
%  \item We evaluate the quality of LLM-generated texts for two tasks.
%  \item We evaluate the ability to discern LLM-generated texts.
%  \item RLHF makes LLM texts lengthier, more repetitive, and detectable.
% \end{highlights}

% Keywords

\maketitle

\section{Introduction}

\gls{llm} \citep{kasneci2023chatgpt}, such as ChatGPT\footnote{https://openai.com/blog/chatgpt/} and deepseek,\footnote{https://www.deepseek.com/} have been widely used in various applications and have been shown to be capable of handling many complex tasks with remarkable performance, such as text summarization \citep{van2024adapted}, question answering \citep{singhal2023towards} and dialogue generation \citep{zhao2023chat}. In these tasks, \gls{llm} demonstrate a text generation ability comparable to, and that can often be perceived as, text written by humans \citep{jakesch2023human}. However, it also raises concerns about the potential for malicious use of LLM-generated texts, given the ease of generating texts at scale which can be harmful, deceitful or inaccurate. The concerns are primarily twofold. First, \gls{llm} are not always trained on up-to-date data, are susceptible to fabrications, which can lead to biased or harmful results, misinformation, and security risks \citep{wu2023survey}. Second, the rapid development of \gls{llm} raises concerns about their potential for malicious exploitation, including for generating harmful content such as deepfakes, fake news, and facilitating academic misconduct \citep{mubarak2023survey}. Therefore, it is essential to develop effective methods to distinguish between human-written and LLM-generated text to avoid misleading users of those texts and to ensure that LLMs are used responsibly.

With the aim of helping determine when a text is generated by LLMs, researchers have developed automated methods to detect human-written vs LLM-generated texts \citep{frohling2021feature,wang2024ai,chakraborty2023possibilities,wang2023seqxgpt}. To date, these include three types of detectors \citep{yang2023survey}: (i) training-based methods, which are trained on labeled samples of human-written and LLM-generated texts and which can perform reasonably well when it encounters new texts that resemble those seen during training; (ii) zero-shot methods, which in the absence of substantial training data look for inherent characteristics of both types of texts to distinguish between them; and (iii) watermarking-based methods, where algorithmically detectable patterns, such as preferred vocabulary words, are added to the generated text as an indicator to facilitate subsequent detection; the latter are however dependent on the use of watermarks at the time of generation, which malicious users can circumvent, and hence in this study we look at the former two methods. Such methods have shown promising results in the detection of LLM-generated text, providing solutions to identify LLM-generated content while mitigating the risks associated with it.

However, the effectiveness of these methods can be challenged by adversarial users who intend to evade detection through the use of techniques to further edit texts to bypass detection. One such technique is \gls{rlhf} \citep{bai2022training}, through which \gls{llm} can be trained to align with human preferences by using human feedback to guide the reinforcement learning process. \gls{rlhf} has also raised concerns about the potential risks and challenges it poses to artificial intelligence (AI) systems. For example, \gls{rlhf} may introduce biases into AI models imposed by its optimization method \citep{xiao2024algorithmicbiasaligninglarge}. Moreover, \gls{rlhf} could exacerbate errors produced by \gls{llm} as it can persuade humans to receive higher rewards from human evaluators \citep{wen2024language}. Reward poisoning attack is another potential risk of \gls{rlhf}, where attackers can manipulate the ranking score to generate texts for malicious purposes \citep{wang2024rlhfpoison}. Hence, detecting LLM-generated text refined with \gls{rlhf} is crucial to ensure the trustworthiness and security of LLM-generated content.

Where \gls{rlhf} can help \gls{llm} generate human-favored outputs, little is known about its impact on blurring the line between human-generated content and LLM-generated text. Thus, we explore how \gls{rlhf} impacts the robustness of current LLM-generated text detection methods to ensure the effectiveness of these methods in detecting LLM-generated text. We train \gls{llm} with \gls{rlhf} on two tasks: question answering and instruction following, and evaluate the performance of the trained models. We then use these models to get LLM-generated texts and also collect human-written texts, which we combine to create a dataset. We feed the dataset into two types of detectors: training-based and zero-shot, and evaluate the performance of the detectors to explore how \gls{rlhf} impacts the robustness of current \gls{llm}-generated text detection methods. Our study leads to the following key findings:

\begin{itemize} 
    \item The outputs of \gls{rlhf}-enhanced \gls{llm} tend to be more detailed and structured, but also lengthier and more repetitive, with decreased syntactic and semantic diversity.
    \item LLM-generated texts refined with \gls{rlhf} tend to be more easily detectable, due to their lengthy and repetitive nature.
    \item The training-based detector is vulnerable to short outputs and mixed texts (e.g. natural language with code snippets), while the zero-shot detector is more robust.
\end{itemize}

\begin{figure*}[ht!]
    \centering
    \includegraphics[width=1\linewidth]{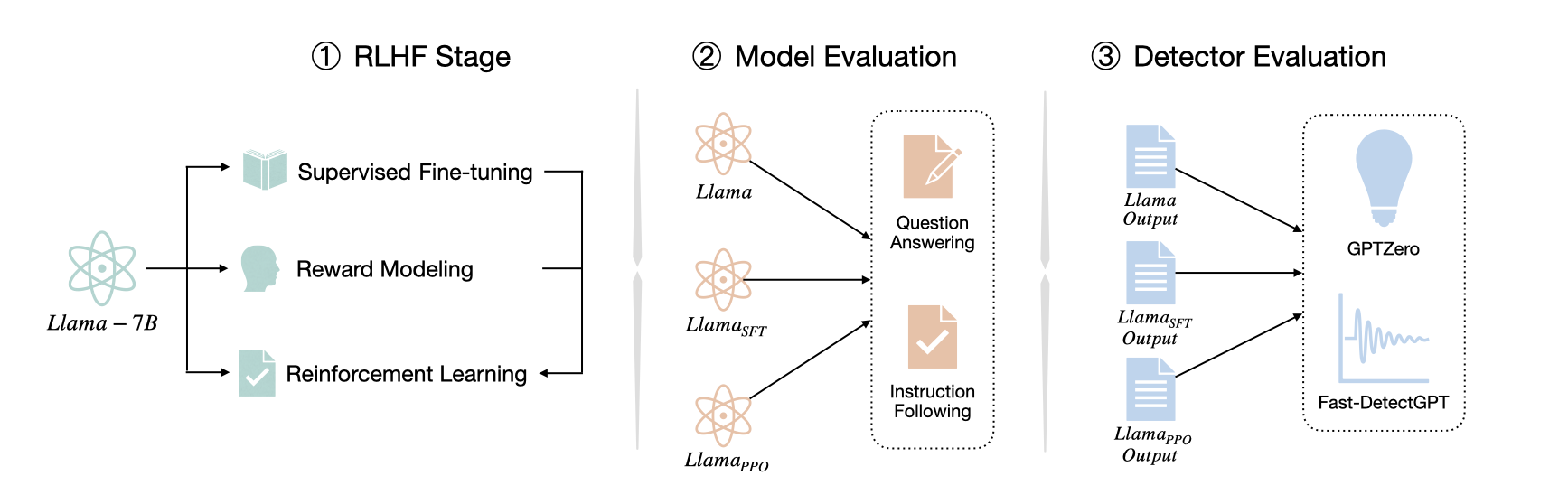}
    \caption{The pipeline. This project contains three main parts: training \gls{llm} with \gls{rlhf}, evaluating the performance of the trained models, and evaluating the performance of different types of detectors.}
    \label{fig:pipeline}
\end{figure*}

The remainder of the paper is organized as follows. We formulate the overall objective and methodology of this research in Section \ref{sec:problem}. Details of training \gls{llm} with \gls{rlhf} on two downstream tasks are presented in Section \ref{sec:training} and Section \ref{sec:tasks}; the AI-generated text detectors for this analysis are introduced in Section \ref{sec:detector}. Evaluation of the trained models as well as different types of detectors is presented in Section \ref{sec:evaluation}. The results and analysis are presented in Section \ref{sec:results}. We conclude the paper in Section \ref{sec:conclusion}.

\section{Related Work} 

In this section, we contextualize our work in the literature by discussing research in three key directions: Large Language Models (LLMs), Reinforcement Learning from Human Feedback (RLHF) and LLM-generated text detection.

\subsection{Large Language Models (LLMs)}

Large Language Models (LLMs) have significantly advanced the field of natural language processing \citep{zhao2023survey,zubiaga2024natural}, enabling a wide range of applications such as text generation, summarization, and question answering. Recent progress in large language models (LLMs) have led to notable improvements in natural language processing tasks, including enhanced reasoning capabilities and performance on complex benchmarks. Moreover, the integration of multimodal inputs, such as text, images, and audio, has expanded the applicability of LLMs across modes \citep{zhang2024mmllmsrecentadvancesmultimodal}. However, the deployment of \gls{llm} requires taking into account ethical considerations \citep{yan2024practical}, including potential biases \citep{ferrara2023should}, hallucinations \citep{rawte2023troubling}, offensive content \citep{weidinger2021ethical} and misleading information \citep{hou2024large} that can be generated as a result. While LLMs have substantially furthered text generation capability, they have also brought about the potential for malicious use \citep{zhao2024silent}, which has motivated research in developing automated methods to detect LLM-generated content (see Section \ref{ssec:llm-gtd}).

\subsection{Reinforcement Learning from Human Feedback (RLHF)}

Reinforcement Learning from Human Feedback (RLHF) is a pivotal technique that has enhanced the alignment of \gls{llm} with human preferences, offering the possibility of improving the quality and reliability of the generated outputs \citep{bai2022training}. The \gls{rlhf} process typically involves three key stages: supervised fine-tuning (SFT), reward modeling (RM), and reinforcement learning (RL), each contributing to the model's ability to produce outputs that are not only more accurate but also aligned with human values \citep{ziegler2019fine,christiano2017deep}. \gls{rlhf} has been instrumental in enhancing the performance of AI in areas where human judgement is crucial. Notable applications include: (i) Natural Language Processing (NLP): improving the quality and safety of language models by aligning generated text with human expectations and reducing harmful or biased outputs \citep{dai2023safe}; (ii) Robotics and Autonomous Systems: enhancing the safety and reliability of autonomous driving systems by incorporating human feedback into the learning process \citep{sun2024optimizing}; (iii) Game Playing Agents: achieving superhuman performance in games by training AI agents with human feedback \citep{ibarz2018rewardlearninghumanpreferences}. Although \gls{rlhf} has been shown to significantly improve the performance of \gls{llm} on various tasks, it is also vulnerable to poisoning attacks \citep{wang2024rlhfpoisonrewardpoisoningattack,pathmanathan2024poisoning} that can induce biases and generate harmful outputs. As such, understanding the impact of \gls{rlhf} on the detectability of \gls{llm}-generated text is crucial to ensure the reliability of detection methods.

\subsection{LLM-Generated Text Detection}
\label{ssec:llm-gtd}

Improvements in LLMs have increased the need for effective LLM-generated text detection methods. These methods are critical in various domains, including misinformation detection, academic integrity, and cybersecurity \citep{mubarak2023survey}. Existing detection approaches can be broadly classified into training-based methods \citep{li2023deepfake}, zero-shot methods \citep{mitchell2023detectgpt}, and watermarking techniques \citep{kirchenbauer2023watermark}. Training-based methods, such as GPTZero, rely on pre-trained models that learn to identify patterns specific to LLM-generated text through supervised learning. On the other hand, zero-shot methods, such as Fast-DetectGPT \citep{bao2023fast}, leverage inherent statistical differences between human- and machine-generated texts without requiring extensive labeled data for training. Watermarking is a technique that adds algorithmically detectable patterns to the generated text as an indicator. However, as \gls{llm} become more adept at mimicking human writing, the detection task may become increasingly complex. For instance, the recent study on the HUMPA (Humanized Proxy Attack) strategy demonstrates that by leveraging a reinforcement learning fine-tuned small language model, attackers can manipulate LLMs to generate text that closely mimics human writing, thereby effectively deceiving state-of-the-art detectors \citep{wang2025humanizingmachineproxyattacks}; DIPPER \citep{krishna2023paraphrasingevadesdetectorsaigenerated} exposed the vulnerability of current AI-generated text detectors when confronted with paraphrasing attacks. This research specifically examines the interaction between \gls{rlhf} and detection methods, assessing whether \gls{rlhf}-induced texts open up new challenges or opportunities for detection. The findings of this study are expected to contribute to the development of more robust detection techniques capable of keeping pace with the rapid evolution of \gls{llm}.

Although significant progress has been made in developing LLM-generated text detection methods, the rapid evolution of \gls{llm}, especially those fine-tuned with \gls{rlhf}, presents unique challenges. Existing detection techniques might struggle to adapt to the nuanced and human-like text outputs produced by \gls{rlhf}-enhanced models. Despite these advancements, there remains a gap in understanding how \gls{rlhf} impacts the detectability of LLM-generated content. This study addresses this critical gap by systematically examining the suitability of current detection methods for \gls{rlhf}-induced texts. Through this analysis, we identify both the challenges and opportunities \gls{rlhf} presents for advancing robust and adaptive detection methodologies.

\section{Problem Formulation} \label{sec:problem}
The primary objective of this research is to investigate how \gls{rlhf} impacts the robustness of current LLM-generated text detection methods. To study this, we first train \gls{llm} with \gls{rlhf} on two tasks: question answering and instruction following. Then, we evaluate the performance of the trained models, and extract model outputs as LLM-generated text. We also collect human-written text to create the dataset and feed it into two types of detectors: training-based and zero-shot to get outputs of detectors. Finally, we evaluate the performance of detectors to assess how \gls{rlhf} impacts the robustness of current \gls{llm}-generated text detection methods. 

\begin{table*}[htb]
    \centering
    % \small
    \caption{Examples from the Question Answering and Instruction Following datasets.}
    \label{tab:data-example}
    \begin{tabular}{|p{0.95\linewidth}|}
    \hline
    \multicolumn{1}{|c|}{\textbf{StackExchange}} \\ \hline
    
    \textbf{Question:} <p>What would be a good tag to use for doing business/making profit with a 3D printer?</p> \\ \hline
    
    \textbf{Answer:} <p>I'd suggest two different tags:</p>
    <ul>
    <li><p>[monetization]: For selling, or profiting from printing, or from printing machines</p></li>
    <li><p>[financing] (or [costs]): For calculating the cost of materials and machines, including operational expenses.</p></li>
    </ul> \\ \hline

    \hline
    \hline

    \textbf{Question:} I have a `\textless{}div id="content"\textgreater{}`. I want to load the content from \textless{}http://vietduc24h.com\textgreater into my `div`: ``` \textless{}html\textgreater \textless{}head\textgreater \textless{}script type="text/javascript"\textgreater{} \$(document).ready(function() \{ \$("\#content").attr("src","http://vietduc24h.com"); \}) \textless{}/script\textgreater \textless{}/head\textgreater \textless{}body\textgreater \textless{}div id="content"\textgreater \textless{}/div\textgreater \textless{}/body\textgreater \textless{}/html ``` I don't want to use an iframe. How can I do this? \\ \hline

    \textbf{Answer:} You need to think about CORS in this aspect. The code you need to have is: ``` \textless{}script type="text/javascript"\textgreater{} \$(document).ready(function() \{ \$("\#content").load("http://vietduc24h.com"); \}) \textless{}/script\textgreater ``` When your domain is not inside `vietduc24h.com`, you might get some security exception. In order to avoid that, you can host a local proxy here. In PHP, we do this way (`url.php`): ``` \textless{}?php \$url = file\_get\_contents(urlencode(\$\_GET{[}"url"{]})); echo \$url; ?\textgreater{} ``` And in the script, you need to modify this way: ``` \textless{}script type="text/javascript"\textgreater{} \$(document).ready(function() \{ \$("\#content").load("proxy.php?url=http://vietduc24h.com"); \}) \textless{}/script\textgreater{} ``` \\ \hline
    
    \hline
    \hline

    \multicolumn{1}{|c|}{\textbf{AlpacaFarm}} \\ \hline
    
    \textbf{Instruction:} Hi, I'm interested in learning to play badminton. Can you explain the game to me? \\ \hline

    \textbf{Response:} Sure! Badminton is a fast-paced racquet sport played by two or four players. The goal is to hit the shuttlecock over the net and into the opponent's side of the court. Points are scored when the other player is unable to return the shuttlecock. You can play singles with one player per side or doubles with two players per side. \\ \hline

    \hline
    \hline

    \textbf{Instruction:} Hi, I'm interested in learning to play badminton. Can you explain the game to me? \\ \hline

    \textbf{Response:} Sure! Badminton is a fast-paced racquet sport played by two or four players. The goal is to hit the shuttlecock over the net and into the opponent's side of the court. Points are scored when the other player is unable to return the shuttlecock. You can play singles with one player per side or doubles with two players per side. \\ \hline
    
    \end{tabular}
\end{table*}

\section{Model Training} \label{sec:training}
The pipeline of our methodology is shown in Figure \ref{fig:pipeline}. In this section, we introduce the training pipeline of \gls{llm} with \gls{rlhf}. Pre-trained \gls{llm} are the base models for the following training steps of supervised fine-tuning, reward modeling, and reinforcement learning. \gls{rlhf} can be processed once the first two steps are completed. 

\subsection{Pre-trained model}
We utilize Llama-7B \citep{touvron2023LLaMA} as our base model, a decoder-only Transformer architecture comprising 7 billion parameters. Llama is an open-source model released by Meta and hence freely available for the research community. We choose this model as it has not been originally fine-tuned using \gls{rlhf}, making it a suitable candidate for our training pipeline.

\subsection{Supervised fine-tuning} 
While \gls{llm} have shown their powerful zero-shot capabilities, they can be further fine-tuned to perform better on specific tasks. We use \gls{sft} to fine-tune the pre-trained model on input-output pairs of downstream tasks using cross-entropy loss. The fine-tuning process aims to improve the model's performance on the specific task by adjusting the model's parameters to minimize the loss between the model's predictions and the ground-truth outputs.

\subsection{Reward modeling} 
We train a reward model following \citet{stiennon2020learning}. Llama is initialized using the \gls{sft} model weights, and then trained given a set of inputs and corresponding outputs with a preference score for each output. The training objective is to predict which output is preferred for a given input and pair of outputs. The reward model is trained by the loss function:
\vspace{-0.1em}

\begin{align}
\text{loss}(\theta) = -\mathbb{E}_{(x, y_j, y_k) \sim D} \Big[ 
    \log \big( \sigma \big( r_{\theta}(x, y_j) - r_{\theta}(x, y_k) \big) \big) \Big] 
\end{align}

\vspace{-0.1em}
where $r$ is the model's score, $\theta$ is model parameters, $\sigma$ is the activation function and $y_j$ is the preferred candidate and $y_k$ is the rejected candidate.

\subsection{Reinforcement learning from human feedback} 
We use \gls{ppo} \citep{schulman2017proximal} as our \gls{rl} algorithm, which is an advanced \gls{rl} algorithm that aims to improve the stability and efficiency of policy updates during training. We also incorporate KL divergence to keep the language model aligned with the \gls{sft} model. Following previous work \citep{pmlr-v70-jaques17a,ziegler2019fine,stiennon2022learningsummarizehumanfeedback}, the final reward for the policy is:
\vspace{-0.1em}
\begin{align}
R(x, y) &= RM_{\theta_{RM}}(x, y) - \beta_{KL} D_{KL}\big(\pi_{\theta_{RL}}(y|x) \parallel \pi_{\theta_{SFT}}(y|x)\big)
\end{align}
\vspace{-0.1em}
where $RM$ is the reward model, $\theta$ denotes the parameters of corresponding models, $\beta_{KL}$ is the KL divergence coefficient, and $D_{KL}$ is the KL divergence between the policy and the \gls{sft} model.

\section{Tasks and Datasets} \label{sec:tasks}
We train \gls{llm} with \gls{rlhf} on two different tasks: question answering and instruction following. We show examples from both datasets in Table \ref{tab:data-example}, which serve as examples to showcase the objectives of both tasks.

\subsection{Question Answering} Question answering (QA) is an NLP task where a model receives questions in natural language, and provides an answer based on a given text or a structured knowledge base. We use the StackExchange dataset \citep{h4stackexchange}, which contains more than 10M question-answer pairs from the Stack Overflow Data Dump\footnote{https://archive.org/details/stackexchange} for model training. The dataset contains general text as well as code (e.g., Python, LaTeX). The dataset was filtered closely following \citet{askell2021generallanguageassistantlaboratory} to ensure that each question has more than two answers, together with the number of upvotes and a label for the accepted answer.

Due to limited computational resources, we use LoRA \citep{hu2021loralowrankadaptationlarge} for parameter-efficient fine-tuning, which drastically reduces the memory footprint. We follow \citet{beeching2023stackLLaMA} for the training pipeline and hyperparameters. We use a 80GB A800 for the training. 

\subsection{Instruction following} Instruction following (IF) is the task where the model receives a direct command or request and generates a response or action that aligns with the given instruction. We use the AlpacaFarm dataset \citep{dubois2024alpacafarm}, where the \texttt{text-davinci-003} model generates the outputs used for the \gls{sft} model, and other \gls{llm} imitate human preferences used to train the reward model. The dataset contains more than 52K instruction-following pairs in total and are divided into four subsets for each step of \gls{rlhf} and validation.

We use SFT, RM and RLHF models released by AlpacaFarm \citep{dubois2024alpacafarmsimulationframeworkmethods}. All of them are trained on the Llama-7B model, and the training objective is to generate responses that are preferred by humans given the instruction.

\section{LLM-generated Text Detector} \label{sec:detector}
To assess the performance of state-of-the-art large language model (LLM) detection techniques on the three text variants we generated--namely, $Llama$, $Llama_{SFT}$, and $Llama_{PPO}$--, we evaluate two categories of detectors: training-based detectors and zero-shot detectors. These detectors are rigorously tested to examine their efficacy in identifying outputs from different model configurations.

\subsection{Training-based detector}
GPTZero\footnote{https://gptzero.me/} is a training-based detector that can distinguish between human-written and LLM-generated text, which has been widely used in academic research \citep{brown2023gptzero,schaaff2024classification,hua2024investigating,zhou2024detecting,de2024text}. It utilizes perplexity and burstiness to determine the origin of a given text since humans tend to write sentences with more variation than machines. Perplexity is measured by the randomness of the text, and burstiness compares sentences with their similarity.

\subsection{Zero-shot detector}
Fast-DetectGPT \citep{bao2023fast} is an effective zero-shot detector that explores the inherent distinctions between human-written and machine-generated texts to identify LLM-generated text. It uses \gls{llm}' conditional probability function for detection and outperforms its baseline DetectGPT \citep{mitchell2023detectgpt} in terms of accuracy and efficiency.

\begin{table}[t]
% \renewcommand{\arraystretch}{1.2}
% \large
\centering
\caption{ROUGE and BERT-Score of models on Question Answering Task}
\label{tab:evaluate-stackLLaMA}
\begin{tabular}{|l|c|c|c|}
\hline
   & \multicolumn{1}{l|}{$Llama$}           & \multicolumn{1}{l|}{$Llama_{SFT}$}      & $Llama_{PPO}$ \\ \hline
  \multicolumn{4}{|c|}{ROUGE-Score}                        \\ \hline
  ROUGE-1 & \textbf{0.1943} & 0.1930          & 0.1793     \\ \hline
  ROUGE-2 & 0.0733          & \textbf{0.0742} & 0.0639     \\ \hline
  ROUGE-L & \textbf{0.1489} & 0.1385          & 0.1322     \\ \hline
  \multicolumn{4}{|c|}{BERT-Score}                         \\ \hline
  Precision & 0.8178 & \textbf{0.8275} & 0.8183 \\ \hline
  Recall    & 0.7932 & \textbf{0.7953} & 0.7909 \\ \hline
  F1-score  & 0.8043 & \textbf{0.8103} & 0.8033 \\ \hline
\end{tabular}
\end{table}

\section{Evaluation of Text Generated by \gls{llm}} \label{sec:evaluation}

Our evaluation is twofold: we first evaluate the quality of the texts generated by the LLMs with and without the SFT and SPO variants, continuing then with the evaluation of the ability of LLM detectors to identify LLM-generated texts under each of the settings.

\subsection{Task 1: Question Answering}

\paragraph{\textbf{Evaluation settings.}} We evaluate our model trained on the Question Answering task based on two aspects: syntactic similarity and semantic similarity. We use ROUGE \citep{lin2004rouge} to measure syntactic similarity; ROUGE (Recall-Oriented Understudy for Gisting Evaluation) is a set of metrics used to evaluate machine-generated text by comparing it to reference texts by measuring n-gram overlap, longest common subsequence (LCS), and skip-bigram similarity. We use ROUGE-1, ROUGE-2, and ROUGE-L as our evaluation metrics. To measure semantic similarity, we use BertScore \citep{zhang2019bertscore}, which computes the similarity between two sentences by comparing the contextual embeddings of the sentences generated by BERT \citep{devlin2019bertpretrainingdeepbidirectional}. We use precision, recall and F1 score as our evaluation metrics.

\paragraph{\textbf{Results.}} Our evaluation results are shown in Table~\ref{tab:evaluate-stackLLaMA}. We can observe that Llama outperforms the other variants in terms of ROUGE-1 and ROUGE-L scores, indicating that it retains a higher degree of unigram overlap and captures the longest common subsequence better than the other models. However, $Llama_{SFT}$ achieves the highest ROUGE-2 score, suggesting that it better captures bigram sequences, which may reflect improved local coherence. The relatively lower performance of $Llama_{PPO}$ for all metrics suggests that, while \gls{rlhf} injects human-aligned preferences into the responses, it also reduces textual overlap with respect to the reference answers, as measured by ROUGE metrics \citep{ye2024uncovering}. When we look at BertScore, $Llama_{SFT}$ outperforms the other models in precision, recall and F1 score, indicating that it generates responses that are semantically more similar to the reference answers.

\subsection{Task 2: Instruction Following}

\paragraph{\textbf{Evaluation settings.}} We use \texttt{AlpacaEval2.0} \citep{alpaca_eval} as the evaluation metric for \gls{rlhf} models trained on the instruction following task. This is an automatic evaluator for instruction following language models containing a fixed set of 805 instructions from the AlpacaFarm evaluation dataset \citep{alpaca}, which represents user interactions on the Alpaca Web demo. Given each instruction, a reference model \texttt{gpt4-turbo-fn} and the evaluated model generate responses, which are passed to a GPT-4 based simulated human-annotator to mark the preferred response. A win rate is subsequently calculated as the likelihood that the automatic evaluator favors the output of the evaluated model across the 805 instructions.

\begin{figure*}[htb]
    \centering
    \includegraphics[width=1\linewidth]{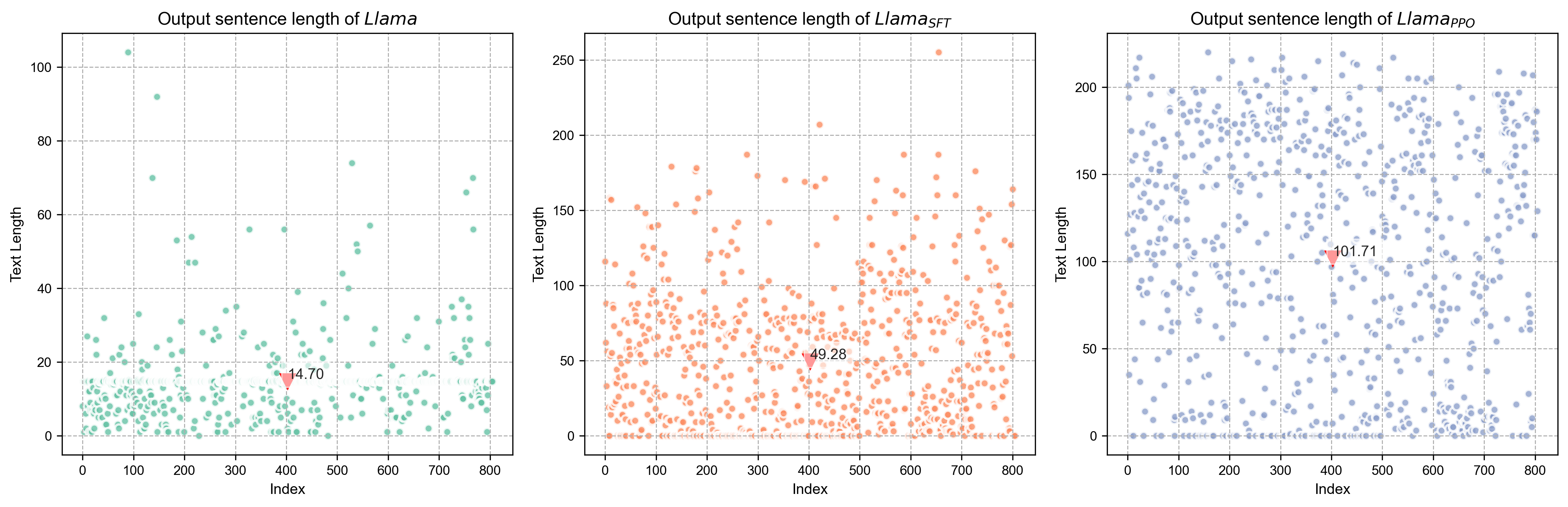}
    \caption{Output sentence length of models. The x-axis represents the index of input, and the y-axis represents the average sentence length of the outputs. The number inside the bar represents the average sentence length of the output.}
    \label{fig:sentence_length}
\end{figure*}

\paragraph{\textbf{Results.}} In Table~\ref{tab:llm-outputs}, we show an example comparing outputs generated by different \gls{llm} for the same input; for more examples of outputs, please refer to Appendix \ref{app:outputs}. We can observe that Llama tends to generate very short responses, which could even be meaningless. The readability of the outputs is significantly improved after \gls{sft}, with more details provided. \gls{rlhf} further improved the responses, with more information and structured sentences. The evaluation results (win rates) of different models are listed in Table~\ref{tab:instruction following evaluation}, where we observe that the quality of the output is significantly improved after \gls{sft} and \gls{rlhf}. To further investigate the impact of \gls{sft} and \gls{rlhf} on the quality of the output, we plot the sentence length of the outputs generated by different models in Figure~\ref{fig:sentence_length}. We can observe that after \gls{sft}, the average sentence length of the outputs is three times longer than that of the outputs generated by Llama in a zero-shot setting. The average sentence length is further increased after \gls{rlhf}, indicating that with further tuning, the model tends to generate longer responses.

\begin{table*}[htbp]
    % \footnotesize
    \caption{Output generated by different \gls{llm}.}
    \label{tab:llm-outputs-1}
    \begin{tabular}{|p{0.15\linewidth}|p{0.80\linewidth}|}
        \hline
        \textbf{Input} & What type of soil is suitable for cactus? \\
        \hline
        \textbf{Reference answer} & Cacti thrive in well-draining soil with high mineral content. A soil mix of sand, perlite, and peat moss or coconut coir is ideal. Avoid heavy clay soils and those that retain moisture for prolonged periods. \\ 
        \hline
        \hline
        \textbf{$Llama$} & Sandy soil is suitable for cactus. \\
        \hline
        \textbf{$Llama_{SFT}$} & Soil suitable for cactus is a well-drained, sandy loam with a pH between 6.5 and 7.5. It should also contain plenty of organic matter, such as compost or manure, to provide nutrients for the cactus. \\
        \hline
        \textbf{$Llama_{PPO}$} & Soil suitable for cactus should be well-drained and sandy, with a pH between 6.5 and 7.5. It should also contain plenty of organic material, such as compost or manure, to provide nutrients for the cactus. The soil should be allowed to dry out completely before more water is added, and the cactus should be watered only when the top few inches of soil feel dry to the touch. Cacti should also be given plenty of sunlight, as they require more direct sunlight than other plants. \\
        \hline
    \end{tabular}
\end{table*}

\section{Evaluation of LLM-generated Text Detection} \label{sec:results}

\paragraph{\textbf{Dataset.}} To evaluate the robustness of the detectors, we expand the datasets by adding human-written text that the models will need to distinguished from the above LLM-generated samples. For the question answering task, we use the ground truth responses of the StackExchange dataset; for the instruction following task, we extract the long-form answers written by human experts from the PubMedQA dataset \citep{jin2019pubmedqa}, which is a large-scale dataset for biomedical question answering. The testing data are sampled from model outputs and human-written texts, respectively. The numbers of samples are balanced between the two classes (100 for each).

\paragraph{\textbf{LLM-generated text detection.}} To evaluate the performance of the detectors, we feed the testing data into the two detectors. The area under the receiver operating characteristic curve (AUROC) is used to evaluate the effectiveness of the detector. This metric quantifies the likelihood that a classifier accurately assigns a higher rank to a randomly chosen positive example (generated by a machine) compared to a randomly chosen negative example (written by a human).

\begin{table}[htbp]
    % \small
    % \renewcommand{\arraystretch}{1.2} 

    \caption{Win rates of different models on the instruction following task.}
    \label{tab:instruction following evaluation}
    \centering
    \begin{tabular}{|l|c|c|c|}
        \hline
        & \textbf{$Llama$} & \textbf{$Llama_{SFT}$} & \textbf{$Llama_{PPO}$} \\
        \hline
        \textbf{win rate} & 1.14 & 9.44 & 20.62 \\
        \hline
    \end{tabular}
\end{table}

\subsection{Comparing performance of LLM-generated text detectors}
We present the performance comparison of different detectors in Table~\ref{tab:performance_comparison}. We can observe that:

\begin{itemize}
    \item Fast-DetectGPT outperforms GPT-Zero overall, indicating that zero-shot detectors are more effective in detecting LLM-generated text.
    \item In the instruction following task, where the input and output data are general text, detectors show a tendency to improve performance when the model is trained with \gls{sft} and \gls{rlhf}. This trend suggests that the iterative training process promotes certain linguistic patterns or stylistic features more commonly seen in LLM-generated texts, decreasing the diversity of the output \citep{kirk2024understanding}, consequently making the outputs more easily detectable. One possible explanation for this phenomenon is that the model, during these training stages, is exposed to a significant amount of LLM-generated texts. Consequently, the model could internalize and reproduce features that are inherently associated with LLM generation, such as specific syntactic structures, lexical choices, or fluency patterns. As a result, the outputs become more distinguishable by LLM detectors.
    \item The performance of the detectors varies across different tasks, and the zero-shot detector is more robust to the question answering task, which contains a mixture of natural language and code snippets. This could be due to the inherent differences between the two detectors. GPT-Zero is a training-based detector, which relies more on its training data. For example, we expect GPT-Zero to be less prepared to deal with code snippets compared to general text detection due to the lack of training data and the complexity of the task \citep{ye2024uncovering}. On the other hand, Fast-DetectGPT is a zero-shot based detector, which pays attention to the inherent differences between human-written and LLM-generated texts. Therefore, this makes Fast-DetectGPT more robust when dealing with the provided mixture of natural language and code snippets.
\end{itemize}

\begin{figure*}[htb]
    \centering
    \includegraphics[width=1\linewidth]{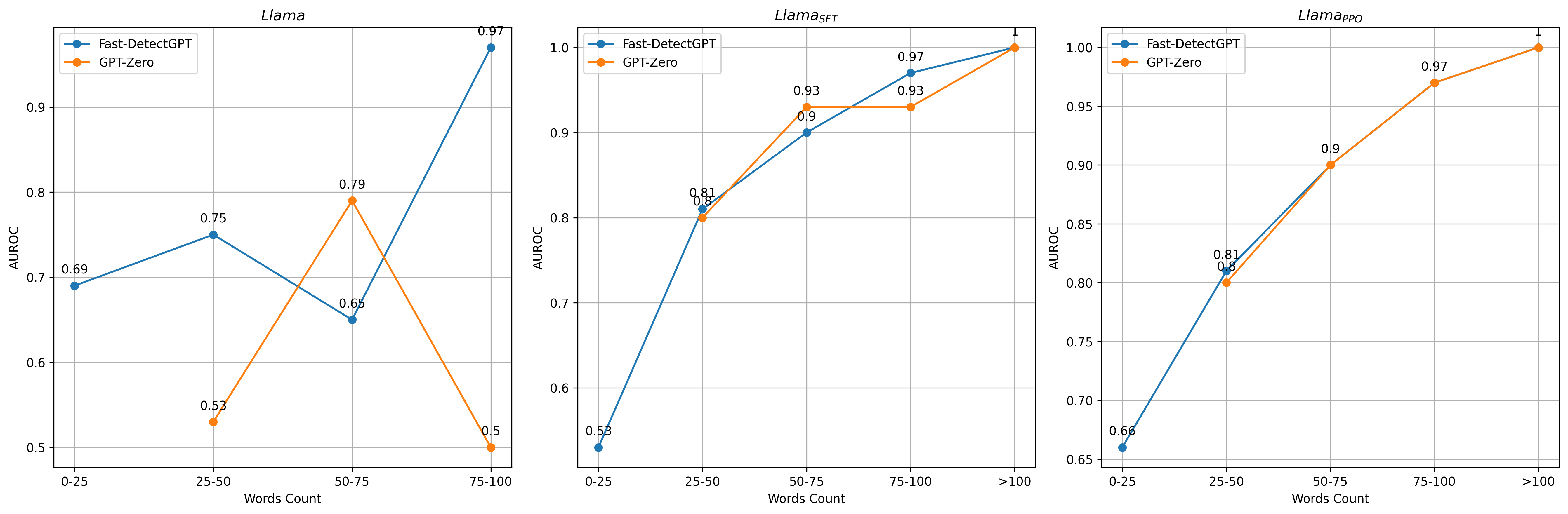}
    \caption{AUROC for detecting samples of different sentence length. The words count of (0-25) is not available for GPT-Zero since it requires the minimum input length of 150 characters. The word count of (>100) is considered in $Llama_{SFT}$ and $Llama_{PPO}$ since there are abundant outputs in these two models.}
    \label{fig:AUROC-sentence_length}
\end{figure*}

\begin{table}[htbp]
  % \small
  \centering
  \caption{AUROC performance for LLM-generated text detection across Llama base model, with SFT and with PPO. FDGPT = Fast-DetectGPT.}
  \label{tab:performance_comparison}
    \begin{tabular}{|l|c|c|c|}
      \hline
      \multicolumn{4}{|c|}{\textbf{Question Answering}} \\ \hline
       & \textbf{$Llama$} & \textbf{$Llama_{SFT}$} & \textbf{$Llama_{PPO}$} \\ \hline
      FDGPT & \textbf{0.91}           & 0.89                & 0.84                \\ \hline
      GPT-Zero       & 0.71                    & \textbf{0.82}       & 0.72                \\ \hline
      \hline
      \multicolumn{4}{|c|}{\textbf{Instruction Following}} \\ \hline
       & \textbf{$Llama$} & \textbf{$Llama_{SFT}$} & \textbf{$Llama_{PPO}$} \\ \hline
      FDGPT & 0.68           & 0.80                & \textbf{0.91}                \\ \hline
      GPT-Zero       & 0.59           & 0.70                & \textbf{0.81}                \\ \hline
    \end{tabular}
\end{table}

\subsection{Analysis of LLM-generated text detection}

\paragraph{\textbf{Sentence Length.}} We further analyze the performance of the detectors in the instruction following task. Since the output length varies across different models, we assess the impact of the output length on the performance of the detectors. The results are presented in Figure~\ref{fig:AUROC-sentence_length}. We can observe that, in general, both detectors perform better on longer outputs, indicating that the detectors are more effective in detecting long content. The training-based detector is more sensitive to sentence length, with a significant growth in performance when the outputs become longer, whereas the zero-shot detector is more robust to short outputs, showing a more stable performance across different sentence lengths.

\paragraph{\textbf{Output Diversity.}} We then analyze the output diversity of different models. There are various aspects in measuring the diversity of the outputs such as syntactic diversity and semantic diversity. To measure syntactic diversity, we follow the distinctive n-grammeme metric \citep{li2015diversity}, which calculates the proportion of distinct n-grammemes in the output, and a higher score indicates higher syntactic diversity. To measure semantic diversity, we use SentenceBERT \citep{reimers2019sentence}, which calculates the cosine similarity between the outputs, and a lower score indicates lower similarity and higher semantic diversity. The results are presented in Table~\ref{tab:diversity}. We can observe that the outputs generated by Llama are more diverse in terms of both syntactic and semantic aspects. The diversity of the outputs decreases after \gls{sft} and \gls{rlhf}, indicating that the training process may reduce the diversity of the outputs, which is likely due to the model's tendency to generate more structured and coherent texts, and the preference for certain linguistic patterns or stylistic features.

\begin{table}[htb]
    % \small
    \centering
    \caption{Output diversity of $Llama$, $Llama_{SFT}$ and $Llama_{PPO}$. Higher Distinct N-gram and lower SentenceBERT score imply higher diversity.}
    \label{tab:diversity}
        \begin{tabular}{|l|c|c|c|}
        \hline
                                  & \textbf{$Llama$} & \textbf{$Llama_{SFT}$} & \textbf{$Llama_{PPO}$} \\ \hline
        \textbf{N-grams} & 0.42           & 0.24                & 0.18                \\ \hline
        \textbf{SBERT}     &   0.06             &     0.09                &   0.07                  \\ \hline
        \end{tabular}
\end{table}

\section{Conclusion}
\label{sec:conclusion}
In this study, we explore how \gls{rlhf} impacts the robustness of current LLM-generated text detection models on two tasks: question answering and instruction following. We train Llama-7B with \gls{rlhf} on the question answering task and use public available models for the instruction following task. We measure the performance of the models using simulated annotators and two types of detectors: training-based and zero-shot. We find that \gls{rlhf} significantly improves the performance of \gls{llm} over the zero-shot setting and \gls{sft} on both tasks. However, the outputs of the \gls{rlhf} model are lengthier and more repetitive, making them more easily detectable by both detectors.

Since a training-based detector adapts to the data seen by the model, its performance drops when dealing with texts that diverge from the seen instances, such as short outputs and mixed texts. The zero-shot detector proves to be more robust to these scenarios, indicating that the intrinsic distinctions between human-written and machine-generated texts are more reliable for detection. We also find that the model tends to generate longer responses after \gls{sft} and \gls{rlhf}, which is likely due to the model's exposure to a significant amount of LLM-generated data during training. The outputs generated by the models are less diverse in terms of both syntactic and semantic aspects after \gls{sft} and \gls{rlhf}, possibly due to the model's tendency to generate more structured and coherent responses, and the preference for certain linguistic patterns or stylistic features.

\section*{Limitations}

This work is not without limitations. First, we only evaluated the performance of the detectors on two tasks: question answering and instruction following. It would be interesting to explore the impact of \gls{rlhf} on other tasks, such as text summarization and dialogue generation. Second, we only used two types of detectors: training-based and zero-shot. It would be beneficial to investigate the performance of other detectors, such as adversarial-based detectors and ensemble-based detectors. Third, we only evaluated the performance of the detectors on the outputs generated by \gls{llm}. It would be valuable to explore the performance of the detectors on other types of AI-generated text, such as text generated by GANs and VAEs.

\printcredits

\section*{Acknowledgments}
The authors express their sincere gratitude to Beijing Shengshu Technology Co., Ltd. for providing the GPU resources that were essential for the completion of this research.


\bibliographystyle{cas-model2-names}

\bibliography{cas}

\appendix

\section{Outputs generated by different \gls{llm}}
\label{app:outputs}

Table \ref{tab:llm-outputs} shows examples of outputs produced by different \gls{llm} for each of the inputs provided.

\begin{table*}[htbp]
    \footnotesize
    \caption{Outputs generated by different \gls{llm}}
    \label{tab:llm-outputs}
    \renewcommand{\arraystretch}{1.2} 
    \begin{tabular}{|p{0.18\linewidth}|p{0.14\linewidth}|p{0.18\linewidth}|p{0.18\linewidth}|p{0.20\linewidth}|}
        \hline
        \textbf{Input} & \textbf{$Llama$} & \textbf{$Llama_{SFT}$} & \textbf{$Llama_{OPP}$} & \textbf{Reference} \\ 
        \hline
        \textbf{What type of soil is suitable for cactus?} & Sandy soil is suitable for cactus. & Soil suitable for cactus is a well-drained, sandy loam with a pH between 6.5 and 7.5. It should also contain plenty of organic matter, such as compost or manure, to provide nutrients for the cactus. & Soil suitable for cactus should be well-drained and sandy, with a pH between 6.5 and 7.5. It should also contain plenty of organic material, such as compost or manure, to provide nutrients for the cactus. The soil should be allowed to dry out completely before more water is added, and the cactus should be watered only when the top few inches of soil feel dry to the touch. Cacti should also be given plenty of sunlight, as they require more direct sunlight than other plants. & Cacti thrive in well-draining soil with high mineral content. A soil mix of sand, perlite, and peat moss or coconut coir is ideal. Avoid heavy clay soils and those that retain moisture for prolonged periods. \\ 
        \hline
        \textbf{How can I make bubble solution?} & You can make bubble solution by mixing water and dish soap. & To make bubble solution, mix 2 cups of water with 1 tablespoon of dish soap and 1/3 cup of white vinegar in a large bowl. Stir the mixture until the soap is completely dissolved. Place the mixture in a spray bottle and shake it well to mix. The bubble solution is now ready to use. & To make bubble solution, you will need 2 cups of water, 1 tablespoon of dish soap, and 1 teaspoon of white vinegar. Start by adding the water and dish soap to a small bowl and stirring until the soap is completely dissolved. Then, add the vinegar and stir until the mixture is completely combined. The bubble solution is now ready to use. & You can make bubble solution by mixing together 1/2 cup of dish soap, 1 1/2 cups of water, and 2 teaspoons of sugar until the sugar is completely dissolved. Optionally, you can add a few drops of food coloring for colored bubbles. \\
        \hline
        \textbf{I have my grandfather's antique fountain pen and a bottle of ink, but have never done this before. Can you tell me how to refill the fountain pen?} & Yes, I can tell you how to refill the fountain pen. & To refill your grandfather's antique fountain pen, you will need a bottle of ink and a small funnel. Start by unscrewing the cap of the fountain pen and filling the small funnel halfway with ink. Then, carefully insert the funnel into the opening of the fountain pen. Slowly pour the ink into the pen until it is half full. Finally, screw the cap back onto the fountain pen and start writing! & To refill a fountain pen, start by unscrewing the cap of the pen. Then, remove the old ink cartridge from the pen by pulling on it gently. After that, take the new ink cartridge and insert it into the pen, making sure that it is facing the correct direction. Finally, screw the cap back onto the pen and start writing! & Sure! First, unscrew the barrel from the nib section. Dip the nib into the ink bottle and twist the converter to draw ink into the pen. Clean the nib with a tissue, reattach the barrel, and test the pen on a scrap paper to ensure it is writing properly. \\
        \hline
    \end{tabular}
\end{table*}

\end{document}